\documentclass[review]{elsarticle}
\usepackage{lineno,hyperref,svg,graphicx}
\usepackage{booktabs}
\usepackage{amsmath}
\usepackage{subcaption}
\usepackage{multirow}
\journal{Journal of Expert Systems with Application}

\biboptions{authoryear}

\begin{document}

\begin{frontmatter}

\title{Enhancing Precision of Automated Teller Machines Network Quality Assessment: Machine Learning and Multi Classifier Fusion Approaches}
\author[1]{Alireza Safarzadeh}
\ead{a.safarzadeh@ut.ac.ir}
\author[2]{Mohammad Reza Jamali}
\ead{jamali@pulseware.ir}
\author[1,3]{Behzad Moshiri \corref{correspondingauthor}}
\cortext[correspondingauthor]{Corresponding author}
\ead{Moshiri@ut.ac.ir}

\address[1]{School of Electrical and Computer Engineering, College of Engineering, University of Tehran, Tehran, Iran}
\address[2]{CEO, Pulseware Company, Tehran, Iran}

\address[3]{Department of Electrical and Computer Engineering, University of Waterloo, Waterloo, Canada}

\begin{abstract}
Ensuring reliable ATM services is essential for modern banking, directly impacting customer satisfaction and the operational efficiency of financial institutions. This study introduces a data fusion approach that utilizes multi-classifier fusion techniques, with a special focus on the Stacking Classifier, to enhance the reliability of ATM networks. To address class imbalance, the Synthetic Minority Over-sampling Technique (SMOTE) was applied, enabling balanced learning for both frequent and rare events.

The proposed framework integrates diverse classification models—Random Forest, LightGBM, and CatBoost—within a Stacking Classifier, achieving a dramatic reduction in false alarms from 3.56\% to just 0.71\%, along with an outstanding overall accuracy of 99.29\%. This multi-classifier fusion method synthesizes the strengths of individual models, leading to significant cost savings and improved operational decision-making.

By demonstrating the power of machine learning and data fusion in optimizing ATM status detection, this research provides practical and scalable solutions for financial institutions aiming to enhance their ATM network performance and customer satisfaction.
\end{abstract}


\end{frontmatter}

\sloppy

\section{Introduction} 

Automated Teller Machines (ATMs) are the mainstay of any modern banking system, where customers expect essential services from them day and night. The performance of these machines is therefore not only vital to ensuring customer satisfaction but also to maintain efficiency in operations by financial institutions. The Key Performance Indicators (KPIs) involved are availability, reliability, Mean Time to Failure (MTTF),  Mean Time to Repair (MTTR), among others, which are important in determining the quality and reliability of the entire ATM network to assist the banking managers. Availability is the proportion of time an ATM network remains 'in-service' compared to 'out-of-service', indicating its operational uptime. Reliability, given by the expression $\text{exp}(-t/\text{MMTF})$, is the likelihood that one of the ATMs will operate without failure for some period $\textit {t}$. MTTF is the total in-service time divided by the number of out-of-service occurrences. MTTR stands for the mean time to repair and put back an ATM in service.

Banking managers base decisions on how well these KPIs are forecast. However, some significant decisional limitations may be brought about by errors in the measurement of ATM status—when an ATM is either out of service and the system does not detect so, or, in the case of a false alarm, if it is functioning properly but the method has signaled it as out of order. These may lead to quite unnecessary maintenance interventions, higher operational costs, and reduced machine availability—affecting customer trust and financial performance. This research aims to enhance KPI accuracy by minimizing errors in detecting out-of-service conditions.

\begin{figure*}[b!]
  \centering
  \resizebox{1\textwidth}{!}{\includegraphics{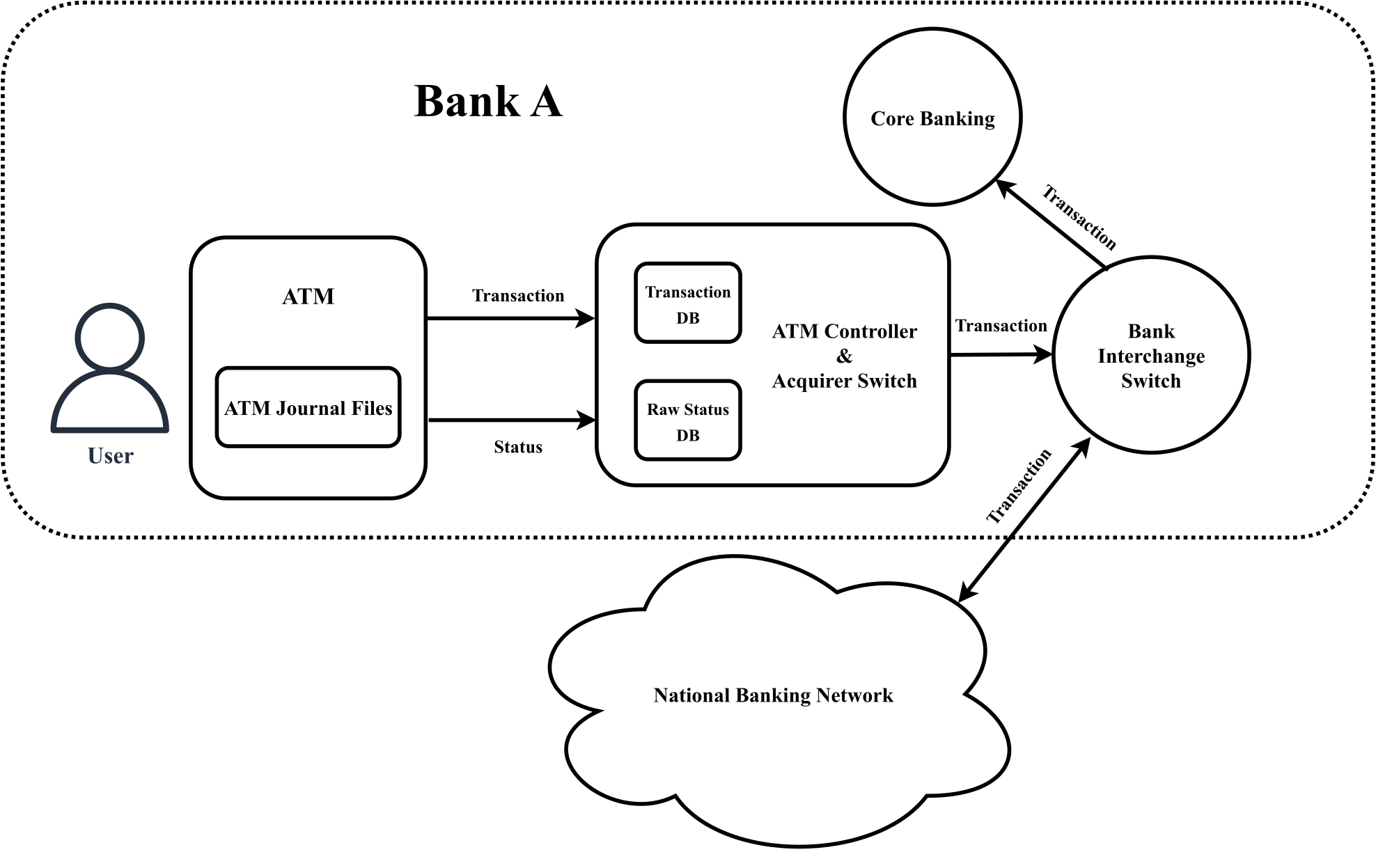}}
  \caption{Schematic representation of the ATM network within the banking system, illustrating data flows between ATMs, status files, transaction records, and the central monitoring system.}
  \label{fig:bank_schema}
\end{figure*}

The detection of out-of-service ATMs could be framed as a classification problem, where every ATM, at an instance in time, can be said to be either in-service or out-of-service. The two major data sources to determine the current state of an ATM are the result of the status files the ATMs push into the Raw Status DB and the ATM Journal Files, where each status change inside the ATM is recorded. Each of these sources, though a couple of examples of the many, feed into the role of the ATM network as part of the larger banking schema illustrated in Figure \ref{fig:bank_schema}. In this context, effectively processing diverse data sources, such as ATM journal files and transaction records, is critical for reliable status detection.

ATM Journal Files provide very verbose logs of detailed actions, right from the moment a user inserts the card into the machine, each and every key pressed, and every single decision made by the ATM itself, such as the selection of banknotes for dispensing. These files are very important in order to understand how an ATM works but usually are not used on a day-to-day basis except for those specific cases when there are disputes, customers report not receiving dispensed cash, among other cases. This is the challenge: to efficiently process these large files for real-time detection of status.

In order to tackle this challenge, the use of financial transaction data is suggested as a complementary indicator of an ATM’s operational status. An ATM that is unable to carry out any financial transaction could very likely be out of service. Therefore, keeping track of the number of transactions performed and their patterns can yield key information on the status of ATMs. In other words, if an ATM is still handling transactions, it can be confirmed as in service, while one that has not recorded any transactions for a significant time period might be classified as out-of-service.

\begin{figure}[b!]
  \centering
  \resizebox{1\textwidth}{!}{\includegraphics{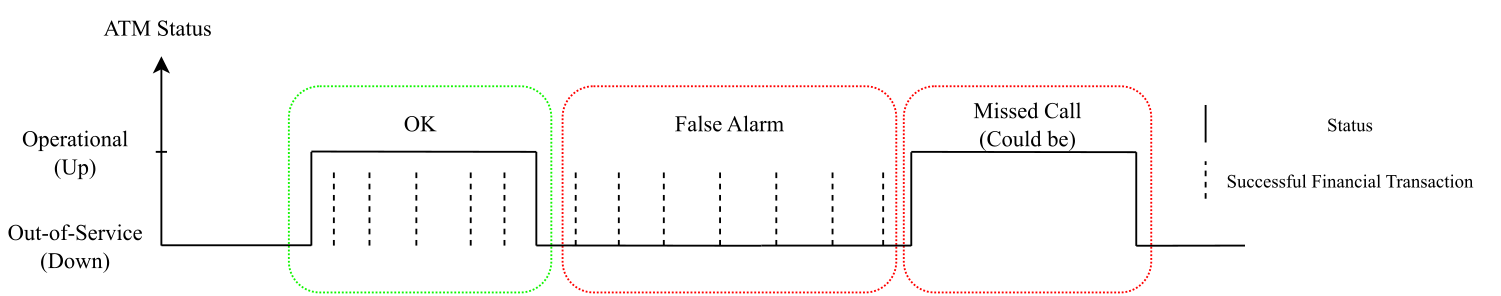}}
  \caption{Example of false alarm and missed alarm scenarios.}
  \label{fig:false_alarm}
\end{figure}

Figure \ref{fig:false_alarm} depicts a situation illustrating both false alarms and missed alarms. To alleviate such issues and enhance ATM reliability, several machine learning models are employed, including Support Vector Machines (SVM), Decision Trees, Ensemble Learning, Dynamic Classifier Selection (DCS), Dynamic Ensemble Selection (DES), Random Forest, stacking Classifier, LightGBM, and CatBoost. These models are adopted as data fusion techniques that combine information from multiple sources, thus boosting the accuracy of ATM status classification. This, in turn, aids banking managers to make more informed decisions with greater precision.

Such problems are alleviated by the use of machine learning models that help improve the reliability of ATMs. These include Support Vector Machines, Decision Trees, Ensemble Learning methods, Dynamic Classifier Selection, Dynamic Ensemble Selection, Random Forest, LightGBM, CatBoost, and Stacking Classifier. Such models are employed due to their data fusion capabilities, where information coming from several sources can be combined, thus serving to increase the accuracy of the status classification of an ATM.
\section{Literature Review}

Since their introduction by Barclays Bank in 1967, ATMs have revolutionized modern banking by enabling self-service for cash withdrawals, balance checks, and fund transfers. Since the first ATM was introduced by Barclays Bank in 1967, these machines have revolutionized banking by enabling self-service for financial transactions \citep{ORegan2018}. ATMs function by reading the customer’s card through a card reader and verifying the provided Personal Identification Number (PIN) entered on the keypad \citep{ORegan2018}.

The core components of an ATM include the card reader, keypad, display, cash dispenser, and receipt printer \citep{ORegan2018}. An ATM is considered “out of service” when it experiences a failure in at least one of the three main areas: a \textit{card reader error}, a \textit{keypad error}, or a \textit{network disconnection}. Such issues prevent the ATM from completing transactions. For instance, a malfunction in the card reader stops the ATM from validating user cards, while a keypad failure prevents PIN entry. A network disconnection renders the ATM incapable of communicating with the bank’s systems, which in turn halts all operations. While issues with other components, such as the cash dispenser, might limit the ATM’s functionality (e.g., it cannot dispense cash), the machine would still be capable of performing other tasks like balance inquiries.

The impact of false alarms, where a system mistakenly indicates an issue, extends beyond banking and can affect multiple industries. Shlomo Breznitz examined the psychological impact of false alarms, noting that repeated false alerts can desensitize operators, making them less likely to respond appropriately to real alarms when they occur \citep{breznitz2013cry}. This phenomenon, termed "crying wolf," is especially problematic in critical systems where missing genuine events could result in significant harm or losses.

Within ATM networks, false alarms trigger unnecessary maintenance activities, driving up operational costs and reducing the availability of machines. For example, a false alarm indicating that an ATM is down can lead to technicians being dispatched to repair a machine that is functioning properly, wasting resources and potentially delaying the resolution of actual issues. Similarly, missed alarms, where a real problem is not detected, prolong downtime, diminish customer satisfaction, and may result in financial penalties for the bank.

Data fusion combines multiple information sources, such as ATM journal files and transaction logs, to produce more reliable conclusions and mitigate issues like false alarms. While traditional probabilistic methods like Bayesian inference have limitations when applied to complex, heterogeneous data, machine learning (ML) provides a robust alternative. ML techniques learn from the data itself, making them particularly effective for modern data fusion tasks \citep{MENG2020115}. For example, Support Vector Machines (SVM) and neural networks enhance data fusion by capturing intricate relationships within noisy or incomplete datasets. Meng et al. (2020) highlighted the effectiveness of both feature-level and decision-level fusion as key areas where machine learning excels. Such techniques have been widely used in fields like autonomous systems and healthcare monitoring, where the integration of multiple data streams is essential for accurate decision-making \citep{MENG2020115}.

Support Vector Machines (SVM) are particularly useful for tackling classification problems, including those involving large-scale data. SVM operates by identifying a hyperplane that best separates the data into distinct classes, efficiently handling both linear and nonlinear separations \citep{Suthaharan2016}. Though computationally demanding, SVM’s robustness makes it suitable for applications across multiple domains.

Decision trees, known for their simplicity and clarity, partition complex decisions into straightforward rules. They are widely adopted in sectors ranging from medical diagnostics to customer analytics \citep{priyanka2020decision}. However, Decision Trees often struggle with overfitting. To address these issues, Random Forests, an ensemble learning approach, were introduced to combine several individual decision trees. Each tree in the forest is built using random subsets of data and features, and their combined predictions reduce the risk of overfitting and improve overall accuracy \citep{breiman2001random}.

While individual models have their strengths, ensemble methods often outperform single classifiers by leveraging diverse perspectives. Ensemble learning methods, including Random Forests, take this concept further by aggregating various models to enhance prediction accuracy. Bagging and boosting techniques are examples of ensemble strategies that compile multiple weak learners to build a stronger predictive model \citep{dong2020survey}.

LightGBM, a highly efficient version of Gradient Boosting Decision Trees (GBDT), improves on traditional models by minimizing computational overhead and memory consumption. Through techniques like Gradient-based One-Side Sampling (GOSS) and Exclusive Feature Bundling (EFB), LightGBM processes large datasets rapidly without sacrificing accuracy \citep{NIPS2017_6449f44a}.

Another GBDT variant, CatBoost, focuses on handling categorical data more effectively. It employs ordered boosting and specialized algorithms for categorical features to mitigate common problems such as target leakage, making it a strong performer on complex data \citep{NEURIPS2018_14491b75}.

Dynamic Classifier Selection (DCS) and Dynamic Ensemble Selection (DES) are techniques used in multi-classifier systems to optimize classifier selection for each instance. Ko et al. (2008) explored these methods in the context of handwritten pattern recognition, demonstrating that dynamically choosing classifiers from an Ensemble of Classifiers (EoC) can significantly improve classification performance \citep{KO20081718}. These techniques inherently rely on the principle of multi-classifier fusion, where diverse models are combined to enhance decision-making.

Unlike static selection, which employs the same set of classifiers for all instances, DCS and DES adaptively pick classifiers based on local accuracy. Ko et al. introduced four new DES schemes utilizing the oracle concept to select an optimal subset of classifiers. Their research showed that these DES methods, coupled with majority voting, outperform static selection by leveraging the variability within classifier ensembles \citep{KO20081718}.

Bagging and boosting methods have turned out to be two of the most popular ensemble learning techniques to improve classification performance. Stacking, introduced by \citep{wolpert1992stacked}, is yet another ensemble technique that can combine multiple classifiers using a meta-learner and improve predictive accuracy on many occasions.

Ensemble learning, particularly stacking methods, has shown great promise in handling complex classification tasks. \citep{KLEIN2023118910} introduced a heterogeneous stacking ensemble combining convolutional neural networks and autoencoders to classify partial discharges in time series data. Their approach demonstrated improved classification accuracy compared to conventional methods, highlighting the effectiveness of leveraging diverse base models in a stacking framework. This work underscores the potential of ensemble methods in scenarios requiring high precision and adaptability, especially when analyzing complex temporal or sequential data.

Stacking ensemble methods have also proven valuable in other fields requiring reliable predictive models. \citep{SAGASTABEITIA2024124930} applied stacking techniques to forecast COVID-19 seroprevalence in the United States. Their study combined diverse methods, including regression models, genetic programming, and neural networks, to enhance model accuracy and generalization. Notably, their results highlight the effectiveness of stacking approaches in integrating complementary strengths of different learning techniques, especially in data-rich environments where diverse features must be synthesized into a cohesive predictive framework.

A common challenge in ML is managing imbalanced datasets, where the minority class (e.g., out-of-service ATMs) is underrepresented compared to the majority class (in-service ATMs). Chawla et al. (2002) developed the Synthetic Minority Over-sampling Technique (SMOTE) to tackle this issue \citep{chawla2002smote}. By generating synthetic examples of the minority class, SMOTE allows the classifier to better recognize underrepresented classes without simply replicating existing instances. In this research, SMOTE was used to balance the dataset, improving recall and precision metrics, particularly for rare events like ATM failures.

Class imbalance is a persistent challenge in machine learning, particularly in classification tasks where one class is underrepresented. Traditional methods like SMOTE have demonstrated effectiveness in improving model performance on imbalanced datasets. However, \citep{KHAN2024122778} emphasize that combining data augmentation techniques, such as SMOTE, with ensemble learning can yield even better results. Their comprehensive evaluation of various methods highlights that traditional augmentation approaches like random oversampling and SMOTE outperform more complex methods like GANs in many scenarios while also being computationally efficient. This underscores the significance of ensemble learning and data augmentation for achieving robust classification outcomes.

Bachmann et al. (2013) demonstrated the advantages of multi-sensor data fusion in freeway traffic speed estimation by combining data from loop detectors and probe vehicles. Their study showed that integrating data from multiple sources yields more accurate results than relying on a single sensor type, a situation comparable to ATM monitoring, where diverse data streams are needed for reliable service detection \citep{bachmann2013comparative}.

Similarly, classifier fusion has been shown to enhance decision-making accuracy. Danesh et al. (2007) proposed a text classification fusion approach that combined Naïve Bayes, k-Nearest Neighbor (k-NN), and Rocchio classifiers using majority voting and OWA operators. Their findings significantly reduced classification errors, illustrating the benefits of fusing diverse classifiers to improve decision accuracy \citep{danesh2007improve}.

Safarzadeh et al. (2021) applied six mathematical models, including Support Vector Machines (SVM) and polynomial functions, to predict COVID-19 trends in Iran before and after the vaccination campaign. The results highlighted that rapid vaccination helped control infection rates and mortality, emphasizing the role of advanced data modeling in tackling real-world problems. While the study focused on public health, its reliance on ML models to interpret data from multiple sources parallels applications in fault detection systems like ATM networks \citep{safarzadeh2021estimating}.

Similarly, false alarms in critical care settings, such as ICUs, highlight the broader challenge of alert fatigue, which is also pertinent in ATM reliability monitoring. Clifford et al. (2016) applied ML algorithms to reduce false arrhythmia alarms, showing that sophisticated models can reduce false alarms by up to 80\% while maintaining sensitivity to genuine events \citep{clifford2016false}.

\section{Methodology}

\subsection{Problem Formulation}
The primary challenge addressed in this study is the frequent occurrence of false alarms in ATM status detection. ATM availability is usually tracked using ATM Status Files, but these files often generate false alerts, leading to unnecessary maintenance activities and misallocation of resources. Conversely, ATM Journal Files offer precise information on ATM status, detailing every operational event and error. However, their substantial size and complexity make real-time processing infeasible.

To address this challenge, the problem was approached as a binary classification task, aiming to categorize each ATM as either in-service or out-of-service. The ground truth labels for this classification were derived from the ATM Journal Files, which underwent extensive preprocessing to serve as the reference standard. Given the significant size constraints of these files, they were used solely for labeling purposes rather than being directly fed into the classification models.

\begin{figure}[b!]
  \centering
  \includegraphics[width=0.85\textwidth]{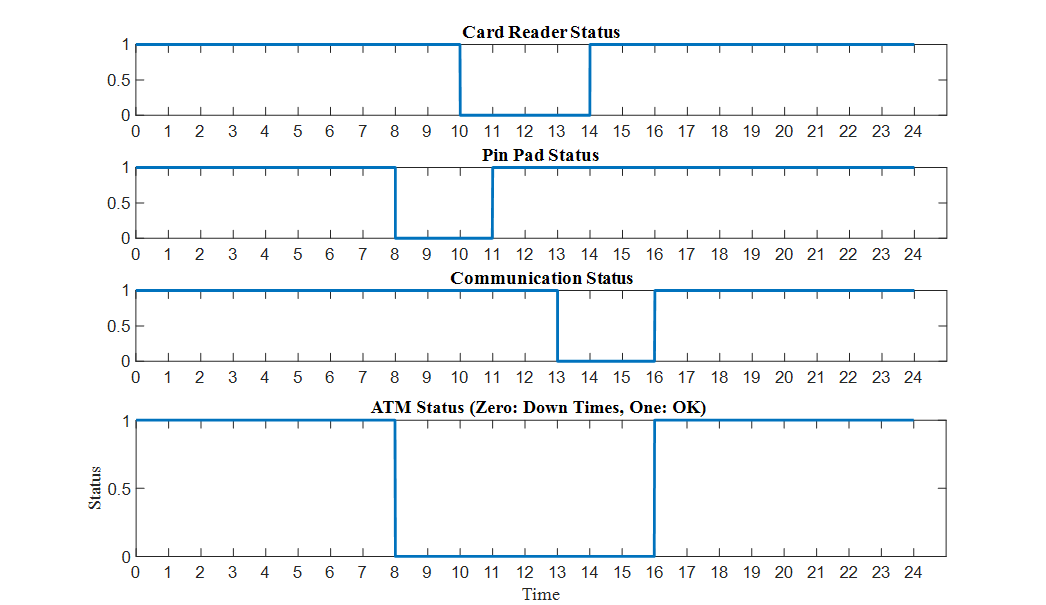}
  \caption{Extraction of out-of-service status from log journal based on errors in card reader, keypad, and communication disconnection.}
  \label{fig:status_extraction}
\end{figure}

Out-of-service statuses were extracted by coding and systematically parsing ATM Journal Files, focusing on component-specific errors such as card reader failures, keypad malfunctions, and network disconnections. Specifically, errors related to the card reader, keypad, and network disconnections were analyzed to identify when an ATM was genuinely out of service. Figure~\ref{fig:status_extraction} illustrates this extraction methodology, highlighting how component-level errors were mapped to the ATM’s operational status.

\subsection{Feature Extraction}
To develop effective classification models, relevant features were extracted from various data sources, primarily focusing on the ATM Status Files and financial transaction records. The following key features were identified and utilized:

\begin{itemize}
    \item \textbf{ATM Status Files}: The operational state of an ATM (either in-service or out-of-service) was derived by analyzing the status of three crucial components: the keypad, card reader, and network connection. If any of these components exhibited errors or if the network connection was interrupted, the ATM was classified as out-of-service. Conversely, an ATM was considered in-service if all components were functioning properly. Figure~\ref{fig:status_extraction} demonstrates the extraction process, showing how component-specific errors are used to determine the overall status of the ATM. Although this data source is susceptible to inaccuracies, including false and missed alarms, it remains a valuable input for modeling. Accounting for these potential inaccuracies is essential in developing a more robust classification strategy for ATM status.

    \item \textbf{Day of the Month}: This feature captures the specific day of the month (ranging from 1 to the final day), providing insights into patterns of ATM usage that may correspond to specific periods, such as salary disbursements or holidays.

    \item \textbf{Type of Working Day}: Days were categorized as regular working days, part-time working days, or holidays. ATMs experience varying demand levels based on the day type, which can influence the likelihood of service disruptions.

    \item \textbf{Time of Day}: The time when an ATM event occurs is crucial for distinguishing between periods of high usage (e.g., midday or evening) and low-traffic periods (e.g., early morning or late at night).

    \item \textbf{Number of Monthly Transactions}: This feature represents the number of financial transactions processed by an ATM within a month, stratified into quantiles. ATMs with higher transaction volumes tend to experience increased wear and tear, making them more prone to service issues.

    \item \textbf{Transaction Status Feature}: Under the assumption that ATM transactions follow a Poisson process, the time intervals between consecutive transactions were modeled using an exponential distribution. The Kolmogorov-Smirnov (KS) test validated this hypothesis, confirming the suitability of the exponential model. This feature provided a supplementary indicator for detecting out-of-service conditions, achieving an accuracy rate of 85.43\%.
\end{itemize}

\subsection{Kolmogorov-Smirnov Test and Probability Threshold}

\begin{figure}[t!]
\centering
\includegraphics[width=0.7\textwidth]{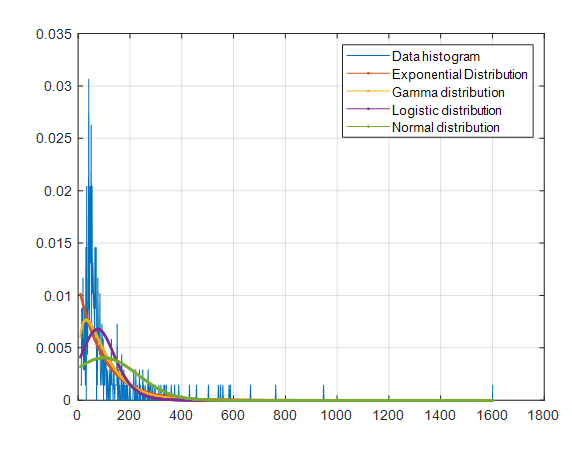}
\caption{Kolmogorov-Smirnov test results for transaction intervals.}
\label{fig:ks_test_figure}
\end{figure}

The Kolmogorov-Smirnov (KS) test was conducted to validate the hypothesis that transaction intervals follow an exponential distribution, a critical assumption for estimating ATM service reliability. Figure~\ref{fig:ks_test_figure} presents the histogram of transaction intervals overlaid with the probability density functions (PDFs) of various distributions, including Exponential, Gamma, Logistic, and Normal. The analysis revealed that the exponential distribution provided the best fit, validating its use for modeling transaction intervals and estimating the likelihood of the next transaction.

\begin{table}[b!]
\centering
\caption{Kolmogorov-Smirnov test results for different distributions.}
\begin{tabular}{@{}ll@{}}
\toprule
\textbf{Distribution} & \textbf{K-S Test Value} \\
\midrule
Exponential & 0.1493 \\
Gamma & 0.1654 \\
Logistic & 0.1906 \\
Normal & 0.2557 \\
\bottomrule
\end{tabular}
\label{table:ks_test_table}
\end{table}

The Kolmogorov-Smirnov (KS) test results, summarized in Table~\ref{table:ks_test_table}, indicate that the exponential distribution provides the best fit for the transaction interval data. Specifically, the KS statistic for the exponential distribution is 0.1493, which is lower than the values for the Gamma (0.1654), Logistic (0.1906), and Normal (0.2557) distributions. These findings support the hypothesis that the intervals between transactions follow an exponential distribution, thereby justifying the use of a Poisson process model to estimate transaction occurrences. This assumption is pivotal because it enables the calculation of the probability of the next transaction occurring within a certain time frame, which is essential for detecting potential out-of-service conditions.

\subsubsection{Probability Threshold Justification}

To identify when an ATM is likely out of service, a 99\% probability threshold was established. If the time elapsed since the last transaction surpasses the 99th percentile of the estimated exponential distribution, it suggests that the ATM is not processing transactions as expected. This high threshold was chosen to strike a balance between sensitivity (correctly identifying out-of-service ATMs) and specificity (minimizing false positives). Setting the threshold at 99\% ensures a high level of confidence before classifying an ATM as potentially out of service, thereby reducing false alarms while still promptly detecting genuine outages.

The KS test validated the assumption that the intervals between consecutive transactions follow an exponential distribution. Using this distribution, the probability of the next transaction is estimated. If this probability exceeds 99\% and no transaction occurs within the expected timeframe, the model assumes that the ATM can no longer process transactions, indicating an out-of-service condition.

\begin{figure}[t!]
  \centering
  \includegraphics[width=0.9\textwidth]{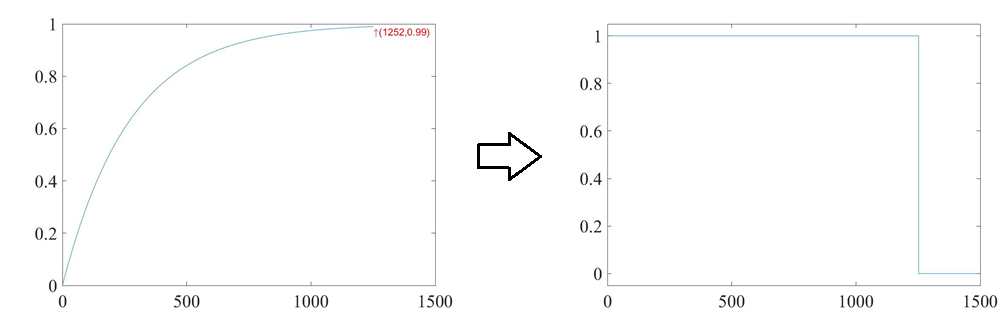}
  \caption{Occurrence of the transaction at moment zero with an average distance between two transactions of 271 seconds (Left) vs. Extracted Status (Right).}
  \label{fig:transaction_occurrence}
\end{figure}

Figure~\ref{fig:transaction_occurrence} illustrates the occurrence of transactions at moment zero, with the average interval between transactions being 271 seconds. This visual demonstrates how the exponential distribution assumption is applied in practice. When no transaction occurs within the estimated interval, the model interprets this as a sign that the ATM has likely gone out of service.

\subsection{Data Imbalance and SMOTE}
The dataset exhibited a significant class imbalance, with only 0.85\% of the instances representing the out-of-service class. To mitigate this issue, the Synthetic Minority Over-sampling Technique (SMOTE) was utilized, with \textit{k = 3} nearest neighbors selected to generate new synthetic instances for the minority class. This approach successfully balanced the dataset to a 50:50 ratio between the in-service and out-of-service classes.

For example, SMOTE generates synthetic samples for out-of-service events, enabling models to better recognize rare instances without overfitting. However, the choice of the \textit{k} parameter is critical to maintain the integrity of the newly generated samples. A smaller \textit{k} value, such as \textit{k=3}, was chosen to ensure that synthetic samples are generated based on the closest neighbors of each minority instance. This method preserves the local structure of the minority class and minimizes the risk of adding noise or outliers to the dataset.

Selecting \textit{k=3} achieves a balance between producing a sufficient number of synthetic samples and maintaining the quality of the data. This parameterization minimizes the potential for overfitting that can occur when synthetic samples are generated using distant neighbors, which may not accurately represent the minority class distribution.

\subsection{Machine Learning Models}
The following machine learning models were applied to the binary classification problem, each trained using the extracted features:

\begin{itemize}
  \item \textbf{Support Vector Machines (SVM)}: A robust classification model, known for its effectiveness in high-dimensional spaces and suitability for binary classification tasks. SVM works by identifying the optimal hyperplane that maximizes the separation margin between different classes.
  
  \item \textbf{Decision Tree}: A straightforward yet powerful model that partitions the feature space into regions based on decision rules, enabling it to handle both categorical and numerical data with ease.

  \item \textbf{Ensemble Learning}:
  \begin{itemize}
    \item \textbf{Bootstrap Aggregation (Bagging)}: An ensemble technique that combines multiple classifiers to reduce variance and improve stability by training each model on randomly sampled subsets of the data.
    
    \item \textbf{Random Forest}: A sophisticated extension of bagging, where decision trees serve as the base learners. It aggregates the predictions from multiple decision trees to reduce overfitting and enhance generalization. Random Forest achieved the highest accuracy in this study, with a score of 99.26\%.
    
    \item \textbf{LightGBM (LGBM)}: A highly efficient Gradient Boosting Decision Tree (GBDT) framework optimized for speed and performance, especially on large datasets. Its leaf-wise tree growth approach makes it a preferred choice for imbalanced datasets.
    
    \item \textbf{CatBoost}: Another GBDT-based algorithm, designed to handle categorical features natively without extensive preprocessing. CatBoost addresses issues like target leakage and reduces overfitting, making it ideal for complex datasets.
    \item \textbf{Stacking Classifier}: Ensemble learning methodology combines several base classifiers for an improved predictive performance by leveraging their complementary strengths. In this study, base learners are Random Forest, LightGBM, and CatBoost, and the meta-learner is Logistic Regression. The stacking classifier is one that combines insights from various algorithms for a more robust model.

  \end{itemize}
  
  \item \textbf{Dynamic Classifier Selection (DCS)} and \textbf{Dynamic Ensemble Selection (DES)}: Advanced ensemble techniques that enhance decision-making by selecting the most suitable classifiers based on the local accuracy for each test instance. These methods ensure that predictions are tailored to the specific characteristics of the data, improving overall classification performance.
\end{itemize}

\subsection{Evaluation Metrics}
Certain key metrics were employed for the performance testing of the classification models for assessing the status of ATM services. The focus was on average scores across the in-service and out-of-service classes for a balanced view of the performance of every model. This includes:

\begin{itemize}

    \item \textbf{Precision}: Precision is the ratio of positively identified instances by the total number of instances predicted as positive. Averaging precision over both classes provides an overall view of the accuracy of each model without weighting any particular class too much. It is given by:
    
    \begin{equation}
        \text{Average Precision} = \frac{\text{Precision}_{\text{down}} + \text{Precision}_{\text{up}}}{2}
    \end{equation}

    This means higher precision reflects the model's ability to minimize false alarms; hence, minimal instances where there was an ATM that had been flagged inappropriately. This improves resource allocation.

    \item \textbf{Recall (Sensitivity)}: Recall is the ratio of correctly predicted positive instances to all actual positive instances. Averaged recall from both classes points to a model's ability to identify in-service and out-of-service status, hence showing balanced performance. It is computed as:
    
    \begin{equation}
        \text{Average Recall} = \frac{\text{Recall}_{\text{down}} + \text{Recall}_{\text{up}}}{2}
    \end{equation}

    High recall signifies that the model detects true instances of each class accordingly, hence giving fewer missed alarms and building reliability.

    \item \textbf{F1-Score}: The F1-score is the harmonic mean of precision and recall and gives a measure that balances the two. Averaging the F1-scores of both classes gives an overall measure of balanced performance. It can be computed as:
    
    \begin{equation}
        \text{Average F1-Score} = \frac{\text{F1-Score}_{\text{down}} + \text{F1-Score}_{\text{up}}}{2}
    \end{equation}

    The average F1-Score shows, quite nicely, the ability of the model to handle both classes well so that decisions are not disproportionately skewed by either false positives or false negatives.

\end{itemize}

Paying attention to average metrics is important in both classes for a number of reasons:

\begin{itemize}
    \item \textbf{Minimizing Operational Disruptions}: This balance between precision in both classes reduces false alarms, missed alarms, optimizes resource utilization, and minimizes superfluous intervention.
    
    \item \textbf{Enhancing Response to Service Status Changes}: High recall across classes ensures timely attention for ATMs that need service, improving customer satisfaction and reducing revenue losses by unresolved service issues.

    \item \textbf{Maximizing Service Reliability}: Any rise in the average F1-score shows that the proposed model is reliable in terms of detecting both in-service and out-of-service status more precisely, hence making for a far more operationally efficient environment and also allowing customers to trust services more.
\end{itemize}

Overall accuracy is generally considered; however, due to issues of class imbalance, it is less informative here. Instead, average metrics—precision, recall, and F1-score—provide a more appropriate, informative estimate of model performance for the reliable management of ATM service status across classes.

\section{Experimental Results}

\subsection{Feature Extraction and Correlation}
The features used in this study include various ATM-related data points such as the day of the month, type of working day, time of day, and the transaction rate of the ATM, which were classified into high or low transaction categories. The transaction rate was normalized into a value ranging from 0 to 1, where ATMs with lower transaction volumes received values closer to 0, and ATMs with higher transaction volumes were assigned values closer to 1. This quantile-based approach helps capture the relationship between transaction load and potential error occurrences.

In addition, the study employs two primary status features: \textit{Extracted Status from ATM Status Files} and \textit{Extracted Status from ATM Transactions}. These features provide binary information on the ATM's operational state, with `1` indicating an in-service status and `0` indicating an out-of-service status.

The type of working day was categorized into five distinct types: first day of the week, last working day, holidays, part-time days, and regular working days. A value between 0 and 1 was assigned to each category based on their frequency of occurrence.

\begin{figure}[t!]
\centering
\includegraphics[width=0.9\textwidth]{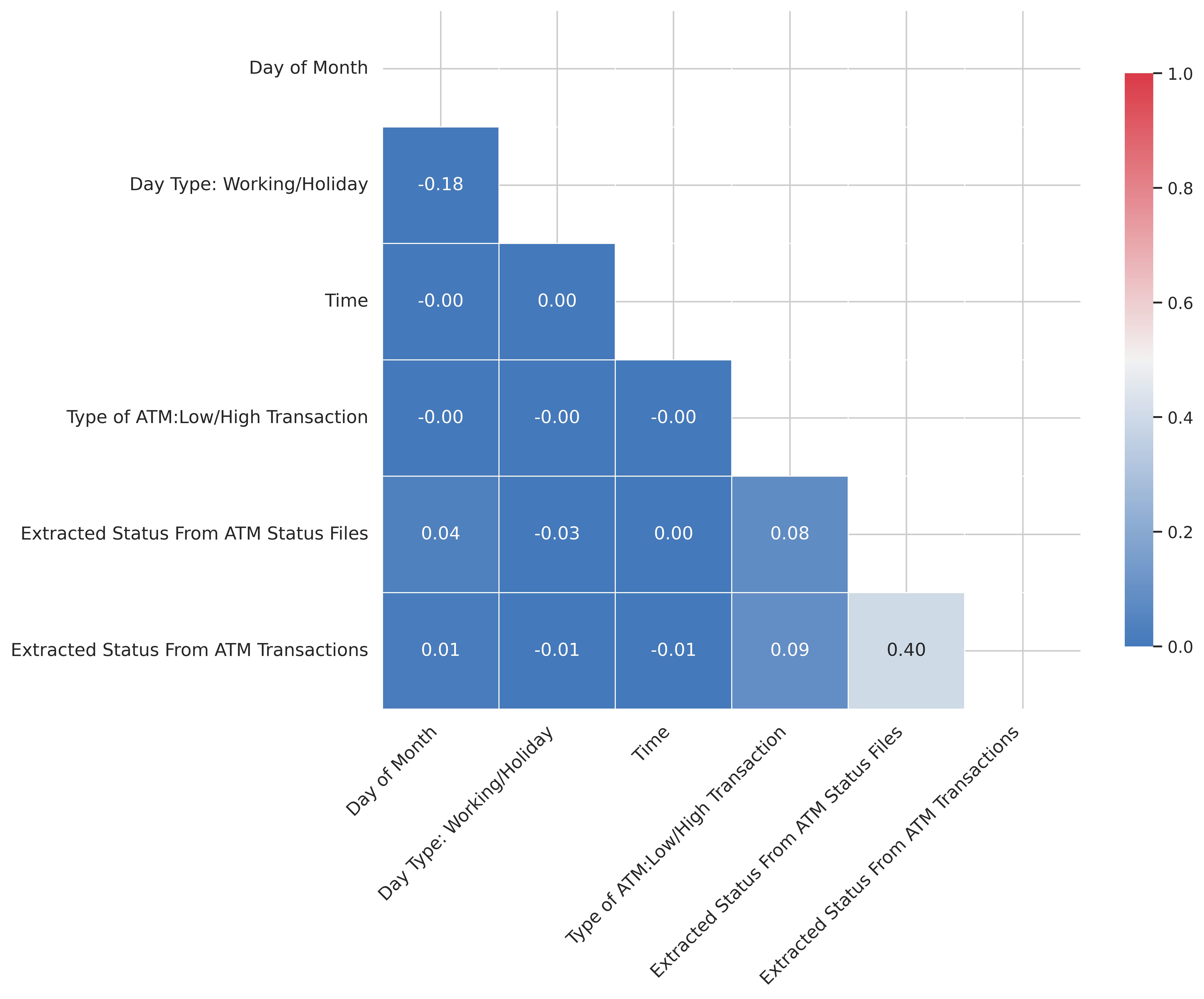}
\caption{Correlation Matrix of Features}
\label{fig:correlation_matrix}
\end{figure}

Figure~\ref{fig:correlation_matrix} illustrates the correlation matrix of these features, showing the relationships and interdependencies between them.

\subsection{ATM Status Files and Extracted Transaction Status Performance}
The ATM Status Files, which indicate the ATM's in-service and out-of-service states, were evaluated using accuracy, precision, recall, and F1-score metrics. Similarly, the statuses derived from the transaction tables, which operate under the assumption that ATM transactions follow a Poisson distribution with exponential inter-arrival times, were also assessed using these metrics.

Table~\ref{table:status_performance} compares the performance of ATM Status Files and transaction log-derived statuses, highlighting the accuracy trade-offs between these sources. The results show that the ATM Status Files achieved an overall accuracy of 96.14\%, while the status extracted from transaction logs resulted in a lower overall accuracy of 85.43\%.

\begin{table}[b!]
\centering
\caption{ATM Status Files and Extracted Transaction Status Performance}
\resizebox{\textwidth}{!}{
\begin{tabular}{|c|c|c|c|}
\hline
\textbf{Metric} & \textbf{Down (Out-of-Service)} & \textbf{Up (In-Service)} & \textbf{Overall Accuracy} \\
\hline
\textbf{ATM Status Files} & & & \\
Precision & 0.1275 & 0.9966 & 0.9614 \\
Recall    & 0.6165 & 0.9643 & \\
F1-Score  & 0.2114 & 0.9802 & \\
\hline
\textbf{Extracted Transaction Status} & & & \\
Precision & 0.0445 & 0.9980 & 0.8543 \\
Recall    & 0.8002 & 0.8548 & \\
F1-Score  & 0.0844 & 0.9209 & \\
\hline
\end{tabular}
}
\label{table:status_performance}
\end{table}

\subsection{Effect of SMOTE on Classification Performance}
To address the class imbalance present in the dataset, the Synthetic Minority Over-sampling Technique (SMOTE) was implemented to upsample the minority class (out-of-service). Initially, the ratio of down to up labels was significantly skewed at 0.0085, indicating a severe imbalance towards in-service (up) labels. After applying SMOTE, the distribution was equalized, resulting in an even representation of in-service and out-of-service instances.

Table~\ref{table:smote_effects} illustrates the impact of SMOTE on the classification performance of three models: Support Vector Machine (SVM), Random Forest, and LightGBM. The precision and recall values for the down (out-of-service) class improved significantly, demonstrating the effectiveness of SMOTE in overcoming the class imbalance issue.

\begin{table}[b!]
\centering
\caption{Effect of SMOTE on Classification Performance}
\resizebox{\textwidth}{!}{
\begin{tabular}{|c|c|c|c|}
\hline
\textbf{Model} & \textbf{Precision (Down)} & \textbf{Recall (Down)} & \textbf{Overall Accuracy} \\
\hline
\textbf{SVM (Before SMOTE)}   & 0.00   & 0.00   & 0.99 \\
\textbf{SVM (After SMOTE)}    & 0.8535 & 0.8747 & 0.8622 \\
\hline
\textbf{Random Forest (Before SMOTE)} & 0.70   & 0.46   & 0.99 \\
\textbf{Random Forest (After SMOTE)}  & 0.9961 & 0.9892 & 0.9926 \\
\hline
\textbf{LightGBM (Before SMOTE)}      & 0.91   & 0.31   & 0.99 \\
\textbf{LightGBM (After SMOTE)}       & 0.9438 & 0.9353 & 0.9398 \\
\hline
\end{tabular}
}
\label{table:smote_effects}
\end{table}

\section{Classification Models and Performance}

\subsection{SVM Classifier}
The Support Vector Machine (SVM) classifier was employed to tackle the binary classification problem of ATM service status prediction. The SVM classifier optimizes the following objective function:

\begin{equation}
\min_{w, b, \zeta} \frac{1}{2} \|w\|^2 + C \sum_{i=1}^{m} \zeta_i
\end{equation}

\[
\text{subject to} \quad y_i \left( w^\top x_i + b \right) \geq 1 - \zeta_i,\quad \zeta_i \geq 0,\quad i = 1, \dots, m
\]

where \( w \) represents the weight vector, \( b \) denotes the bias term, \( C \) is the regularization parameter, and \( \zeta_i \) are slack variables that allow for classification errors. The SVM model works by maximizing the margin between the two classes (in-service and out-of-service) while penalizing any misclassifications.

The following parameters were configured for the SVM model:

\begin{itemize}
    \item \textbf{Penalty Term (C)}: The regularization parameter was set to 1, striking a balance between maximizing the margin and minimizing classification errors.
    \item \textbf{Loss Function}: The squared hinge loss was utilized, defined as:

    \begin{equation}
    \sum_{i=1}^{n} \max(0, 1 - y_i (w^\top x_i + b))^2
    \end{equation}

    This loss function penalizes misclassified samples that lie beyond the decision boundary, ensuring that the model places emphasis on difficult examples.
    
    \item \textbf{Kernel}: A linear kernel was chosen for this study, which is effective for separating the two classes in high-dimensional feature space.
\end{itemize}

The performance of the SVM classifier was evaluated both before and after applying SMOTE. SMOTE significantly enhanced the precision, recall, and F1-score for the down (out-of-service) class.

This improvement underscores the importance of SMOTE in addressing class imbalance, allowing the SVM model to more effectively identify the minority class instances.

Table~\ref{table:svm_performance} presents the performance metrics for the SVM classifier trained with SMOTE.

\subsection{Decision Tree Classifier}
The Decision Tree classifier was applied to solve the classification problem in this study. The model’s parameters are outlined as follows:

\begin{table}[t!]
\centering
\caption{SVM Performance After SMOTE}

\begin{tabular}{|c|c|c|c|}
\hline
\textbf{Metric} & \textbf{Down (Out-of-Service)} & \textbf{Up (In-Service)} & \textbf{Overall Accuracy} \\
\hline
Precision & 0.8535 & 0.8715 & 0.8622 \\
Recall    & 0.8747 & 0.8498 & \\
F1-Score  & 0.8639 & 0.8605 & \\
\hline
\end{tabular}
\label{table:svm_performance}
\end{table}

\begin{itemize}
    \item \textbf{Criterion}: The Gini impurity was used as the splitting criterion, which measures the quality of a split by evaluating the distribution of classes across the branches.
    \item \textbf{Minimum Samples for Split}: A minimum of 2 samples was required to perform a split at each node, ensuring that each decision node has sufficient data to make a reliable split.
    \item \textbf{Minimum Samples for Leaf}: The minimum number of samples in a leaf node was set to 1, which allows the model to grow deeper trees to capture finer details in the data.
\end{itemize}

The Decision Tree model creates branches by recursively splitting the data based on the feature that maximizes the Gini impurity reduction at each step. This process continues until the stopping criteria (minimum samples per split or leaf) are met, resulting in a tree structure that assigns a label to each leaf node, representing the predicted class (either in-service or out-of-service).

Table~\ref{table:decision_tree_metrics} presents the performance metrics of the Decision Tree classifier on the dataset.
\begin{table}[b!]
\centering
\caption{Decision Tree Performance Metrics}
\begin{tabular}{|c|c|c|c|}
\hline
\textbf{Metric} & \textbf{Down (Out-of-Service)} & \textbf{Up (In-Service)} & \textbf{Overall Accuracy} \\
\hline
Precision & 0.9929 & 0.9841 & 0.9884 \\
Recall    & 0.9839 & 0.9929 & - \\
F1-Score  & 0.9884 & 0.9885 & - \\
\hline
\end{tabular}
\label{table:decision_tree_metrics}
\end{table}
\subsection{Bagging Classifier}
The Bagging (Bootstrap Aggregation) classifier was implemented as an ensemble learning approach to enhance prediction accuracy by combining multiple base models. In this study, Decision Trees were used as the base learners for Bagging. This strategy stabilizes individual model predictions, reducing variance and improving overall reliability.

The key parameters used in the Bagging model were as follows:

\begin{itemize}
    \item \textbf{Base Model}: Decision Trees were chosen as the base models due to their high interpretability and ability to capture complex patterns.
    \item \textbf{Number of Base Learners}: A total of three Decision Trees were trained as individual base learners, providing diverse perspectives on the data.
    \item \textbf{Max Samples}: Each base learner was trained using 100\% of the original dataset, ensuring complete representation and consistency.
    \item \textbf{Max Features}: All features (100\%) were used by each base learner, ensuring that each tree had access to the same set of information.
\end{itemize}

Bagging is particularly useful when base models are prone to overfitting, as in the case of Decision Trees, by aggregating predictions across multiple versions to generate a consensus output.

\begin{table}[b!]
\centering
\caption{Bagging Classifier Performance Metrics}
\begin{tabular}{|c|c|c|c|}
\hline
\textbf{Metric} & \textbf{Down (Out-of-Service)} & \textbf{Up (In-Service)} & \textbf{Overall Accuracy} \\
\hline
Precision & 0.9952 & 0.9885 & 0.9918 \\
Recall    & 0.9884 & 0.9953 & - \\
F1-Score  & 0.9918 & 0.9919 & - \\
\hline
\end{tabular}
\label{table:bagging_metrics}
\end{table}

The performance metrics in Table~\ref{table:bagging_metrics} show that the Bagging classifier achieved high precision and recall for both the out-of-service and in-service classes, making it a reliable approach for minimizing false alarms and missed detections.

\subsection{LightGBM Classifier}
The LightGBM classifier, known for its efficiency and scalability, was utilized in this study to classify ATMs as either in-service or out-of-service. LightGBM is particularly effective for large datasets due to its fast training speed and low memory footprint. Additionally, it constructs trees leaf-wise rather than level-wise, enabling faster convergence and better loss reduction. It also handles categorical data effectively, making it a strong candidate for real-world applications.

The model parameters used for the LightGBM classifier were:

\begin{itemize}
    \item \textbf{Number of Leaves}: 31 (sets the maximum number of leaves in a single tree).
    \item \textbf{Max Depth}: -1 (no limit on tree depth to ensure full growth).
    \item \textbf{Number of Estimators}: 100 (indicates the total number of boosting iterations).
    \item \textbf{Random State}: Not specified (controls randomness for reproducibility if set).
\end{itemize}

LightGBM's leaf-wise growth strategy provides it with the ability to reduce more loss at each step, making it computationally more efficient than traditional gradient boosting methods. This characteristic is particularly beneficial when dealing with extensive datasets.

\begin{table}[b!]
\centering
\caption{LightGBM Performance Metrics}
\begin{tabular}{|c|c|c|c|}
\hline
\textbf{Metric} & \textbf{Down (Out-of-Service)} & \textbf{Up (In-Service)} & \textbf{Overall Accuracy} \\
\hline
Precision & 0.9438 & 0.9359 & 0.9398 \\
Recall    & 0.9353 & 0.9443 & - \\
F1-Score  & 0.9396 & 0.9401 & - \\
\hline
\end{tabular}
\label{table:lgbm_metrics}
\end{table}

The performance metrics presented in Table~\ref{table:lgbm_metrics} show that the LightGBM model maintained a high level of precision and recall for both in-service and out-of-service classes, achieving an overall accuracy of 93.98\%. This highlights its robustness and capability in handling the ATM classification task with minimal errors.

\subsection{CatBoost Classifier}
CatBoost, another gradient boosting algorithm, was used to classify ATMs as either in-service or out-of-service. This model is particularly effective at handling categorical data without requiring extensive preprocessing, making it a robust choice for this study. CatBoost, developed by Yandex, is designed to reduce overfitting and perform well on datasets of varying sizes. Its key strength lies in its use of ordered boosting and symmetrical trees, which help minimize prediction bias and maintain model integrity even with complex data.

The CatBoost model was trained using the default parameters provided by the open-source CatBoost library, ensuring a reliable and standard approach for benchmarking its performance against other models.

\begin{table}[b!]
\centering
\caption{CatBoost Performance Metrics}
\begin{tabular}{|c|c|c|c|}
\hline
\textbf{Metric} & \textbf{Down (Out-of-Service)} & \textbf{Up (In-Service)} & \textbf{Overall Accuracy} \\
\hline
Precision & 0.9853 & 0.9843 & 0.9848 \\
Recall    & 0.9843 & 0.9854 & - \\
F1-Score  & 0.9848 & 0.9848 & - \\
\hline
\end{tabular}
\label{table:catboost_metrics}
\end{table}

Table~\ref{table:catboost_metrics} illustrates the performance of CatBoost, demonstrating high precision and recall for both the in-service and out-of-service categories, with an overall accuracy of 98.48\%. This strong performance underscores CatBoost's suitability for handling complex classification tasks, ensuring reliable and consistent results even with limited tuning.

\subsection{Dynamic Classifier Selection (DCS)}
Dynamic Classifier Selection with Local Accuracy (DCS LA) is a technique that dynamically selects the most suitable classifier for each test instance based on its local accuracy within a defined neighborhood of training samples. In this study, a pool of classifiers, comprising Decision Trees and Support Vector Machines (SVM), was used. The DCS LA method evaluates the performance of each classifier in the pool and chooses the one that demonstrates the highest local accuracy for the specific test instance.

The main parameters for configuring the DCS LA model are as follows:

\begin{itemize}
    \item \textbf{Classifier Pool}: The pool includes Decision Trees and Support Vector Machines (SVM).
    \item \textbf{Number of Neighbors}: The local accuracy is determined using the 7 nearest neighbors for each test instance.
\end{itemize}

The performance of the DCS LA method is summarized in Table~\ref{table:dcs_metrics}, highlighting its effectiveness in classifying ATMs into in-service or out-of-service categories.

\begin{table}[b!]
\centering
\caption{DCS LA Performance Metrics}
\begin{tabular}{|c|c|c|c|}
\hline
\textbf{Metric} & \textbf{Down (Out-of-Service)} & \textbf{Up (In-Service)} & \textbf{Overall Accuracy} \\
\hline
Precision & 0.9950 & 0.9890 & 0.9920 \\
Recall    & 0.9889 & 0.9951 & - \\
F1-Score  & 0.9920 & 0.9920 & - \\
\hline
\end{tabular}
\label{table:dcs_metrics}
\end{table}

The results in Table~\ref{table:dcs_metrics} show that DCS LA achieves an overall accuracy of 99.20\%, making it a reliable choice for ATM status classification by dynamically selecting the best-performing classifier based on local accuracy.

\subsection{Dynamic Ensemble Selection (DES)}
Dynamic Ensemble Selection (DES) with K Nearest Oracles Eliminate (KNORAE) is a technique designed to identify the best ensemble of classifiers for a given test instance based on their local performance within a neighborhood of test samples. This method leverages the K nearest neighbors to assess the classifiers’ effectiveness, dynamically choosing the optimal ensemble for each prediction scenario.

The primary parameters for configuring the DES KNORAE model are outlined below:

\begin{itemize}
    \item \textbf{Classifier Pool}: Bagging classifiers were utilized as the base learners to construct the ensembles.
    \item \textbf{Number of Neighbors}: The method considered the 7 nearest neighbors to determine classifier performance for each instance.
    \item \textbf{Voting Mechanism}: Majority voting was employed to aggregate predictions from the chosen ensemble.
\end{itemize}

Table~\ref{table:des_metrics} outlines the performance metrics of the DES KNORAE method, demonstrating its efficiency in ATM status classification.

\begin{table}[b!]
\centering
\caption{DES KNORAE Performance Metrics}
\begin{tabular}{|c|c|c|c|}
\hline
\textbf{Metric} & \textbf{Down (Out-of-Service)} & \textbf{Up (In-Service)} & \textbf{Overall Accuracy} \\
\hline
Precision & 0.9934 & 0.9862 & 0.9898 \\
Recall    & 0.9861 & 0.9935 & - \\
F1-Score  & 0.9897 & 0.9898 & - \\
\hline
\end{tabular}
\label{table:des_metrics}
\end{table}

The results in Table~\ref{table:des_metrics} indicate that DES KNORAE achieves a high precision and recall, making it a robust method for ensuring accurate ATM status prediction through dynamic ensemble selection.

\subsection{Random Forest Classifier}
Random Forest is an ensemble learning technique that builds multiple decision trees (weak learners) and combines their outputs to generate a more robust and accurate prediction. Each tree is constructed using a random subset of the data, and the final decision is based on the majority vote of all the trees in the ensemble, which helps to mitigate overfitting and variance issues commonly found in single decision tree models.

The configuration parameters used for the Random Forest model in this study are as follows:

\begin{itemize}
    \item \textbf{Number of Trees}: A total of 100 decision trees were employed in the forest to ensure sufficient learning capacity.
    \item \textbf{Criterion}: The Gini impurity index was chosen as the splitting criterion to measure the quality of the splits.
    \item \textbf{Minimum Samples per Leaf}: Each leaf node was required to have at least one sample to prevent the model from creating overly complex trees.
\end{itemize}

Table~\ref{table:rf_metrics} provides a summary of the performance metrics for the Random Forest classifier.

\begin{table}[b!]
\centering
\caption{Performance Metrics of the Random Forest Classifier}
\begin{tabular}{|c|c|c|c|}
\hline
\textbf{Metric} & \textbf{Down (Out-of-Service)} & \textbf{Up (In-Service)} & \textbf{Overall Accuracy} \\
\hline
Precision & 0.9961 & 0.9892 & 0.9926 \\
Recall    & 0.9892 & 0.9961 & - \\
F1-Score  & 0.9926 & 0.9927 & - \\
\hline
\end{tabular}
\label{table:rf_metrics}
\end{table}

As seen in Table~\ref{table:rf_metrics}, the Random Forest model demonstrated high precision and recall for both classes, making it a reliable choice for ATM status classification.

\subsection{Stacking Classifier}
The stacking classifier is an ensemble learning method that combines multiple classification models. In this paper, the stacking classifier technique was applied to improve predictive performance by leveraging the complementary strengths of several models \citep{wolpert1992stacked}. In our study, a stacking classifier was implemented to enhance the accuracy of ATM status prediction.
\begin{itemize}
\item \textbf{Base Learners}: The base classifiers in the study were Random Forest, LightGBM, and CatBoost. They were selected based on their excellent individual performances and diversity in learning algorithms. Random Forest is an ensemble of decision trees that reduces variance through bagging. LightGBM is an efficient gradient boosting framework optimized for speed and performance. CatBoost is another gradient boosting algorithm that handles categorical features effectively.
\item \textbf{Meta-Learner}: We used a Logistic Regression model as the meta-learner. Logistic Regression is applicable for binary classification tasks and effectively combines the predictions from various base learners.
\item \textbf{Training Process}:
\begin{enumerate}
    \item \textbf{First Level}: The base learners were trained on the training dataset after the application of SMOTE to handle class imbalance. Each model was trained using their optimized parameters independently.
    \item \textbf{Second Level}: The base learners' predictions were used as input features to the meta-learner. The meta-learner minimized the cross-entropy loss to optimize the final prediction.
\end{enumerate}
\item \textbf{Parameter Settings}:
\begin{itemize}
    \item \textbf{Random Forest}: 100 trees, Gini impurity criterion, random state set to 1.
    \item \textbf{LightGBM}: 100 estimators, learning rate of 0.1, random state set to 1.
    \item \textbf{CatBoost}: 100 iterations, learning rate of 0.1, depth of 6, silent mode enabled.
    \item \textbf{Logistic Regression}: Using default setting with L2 regularization.
\end{itemize}
\end{itemize}
\textbf{Performance Metrics}:
The performance of the stacking classifier is summarised in the following Table~\ref{table:stacking_metrics}.
\begin{table}[t!]
\centering
\caption{Performance Metrics of the Stacking Classifier}
\begin{tabular}{|c|c|c|c|}
\hline
\textbf{Metric} & \textbf{Down (Out-of-Service)} & \textbf{Up (In-Service)} & \textbf{Overall Accuracy} \\
\hline
Precision & 0.9949 & 0.9910 & 0.9929 \\
Recall    & 0.9910 & 0.9949 & - \\
F1-Score  & 0.9929 & 0.9930 & - \\
\hline
\end{tabular}
\label{table:stacking_metrics}
\end{table}
The overall accuracy of the stacking classifier was 99.29\% among all models tested. For both classes, precision and recall were high, which is indicative of its performance in reducing both false alarms and missed alarms.

\section{Model Comparison and Analysis}

This section provides a detailed comparison of the performance of various classification models used in this study. The models are evaluated based on key metrics: accuracy, precision, recall, and F1-scores for both in-service (Up) and out-of-service (Down) classes. Table~\ref{table:comparison_metrics} presents the summarized results, providing a comprehensive overview of each model's strengths and weaknesses.

\begin{table}[b!]
\centering
\caption{Model Performance Comparison for Down (Out-of-Service) and Up (In-Service) Classes}
\resizebox{\textwidth}{!}{
\begin{tabular}{|c|c|c|c|c|c|c|}
\hline
\textbf{Model} & \textbf{Precision (Down)} & \textbf{Recall (Down)} & \textbf{F1-Score (Down)} & \textbf{Precision (Up)} & \textbf{Recall (Up)} & \textbf{F1-Score (Up)} \\
\hline
\textbf{SVM}           & 0.8535 & 0.8747 & 0.8639 & 0.8715 & 0.8498 & 0.8605 \\
\textbf{Decision Tree}  & 0.9929 & 0.9839 & 0.9884 & 0.9841 & 0.9929 & 0.9885 \\
\textbf{Bagging}        & 0.9952 & 0.9884 & 0.9918 & 0.9885 & 0.9953 & 0.9919 \\
\textbf{LightGBM}       & 0.9438 & 0.9353 & 0.9396 & 0.9359 & 0.9443 & 0.9401 \\
\textbf{CatBoost}       & 0.9853 & 0.9843 & 0.9848 & 0.9843 & 0.9854 & 0.9848 \\
\textbf{DCS LA}         & 0.9950 & 0.9889 & 0.9920 & 0.9890 & 0.9951 & 0.9920 \\
\textbf{DES KNORAE}     & 0.9934 & 0.9861 & 0.9897 & 0.9862 & 0.9935 & 0.9898 \\
\textbf{Random Forest}  & 0.9961 & 0.9892 & 0.9926 & 0.9892 & 0.9961 & 0.9927 \\
\textbf{Stacking Classifier}  & 0.9949 & 0.9910 & 0.9929 & 0.9910 & 0.9949 & 0.9930 \\
\hline
\end{tabular}
}
\label{table:comparison_metrics}
\end{table}
While all models appear to be doing quite well, which can be confirmed from Table~\ref{table:comparison_metrics}, the Stacking Classifier slightly outperformed the rest, yielding the highest average F1-scores across both classes, reaching 0.9929 for the class Down (Out-of-Service) and 0.9930 for the class Up (In-Service). Also, Random Forest, DCS LA, and Bagging models tend to yield relatively high values of precision, recall, and F1-score, rather close to those obtained by the Stacking Classifier. Whereas SVM and LightGBM were effective, they didn't reach the same level of precision and recall for both classes. Neither has been outshine by those ones to capture complex patterns in approaches of ensembles nor improve their performance of classification.

\begin{figure}[b!]
\centering
\includegraphics[width=0.9\textwidth]{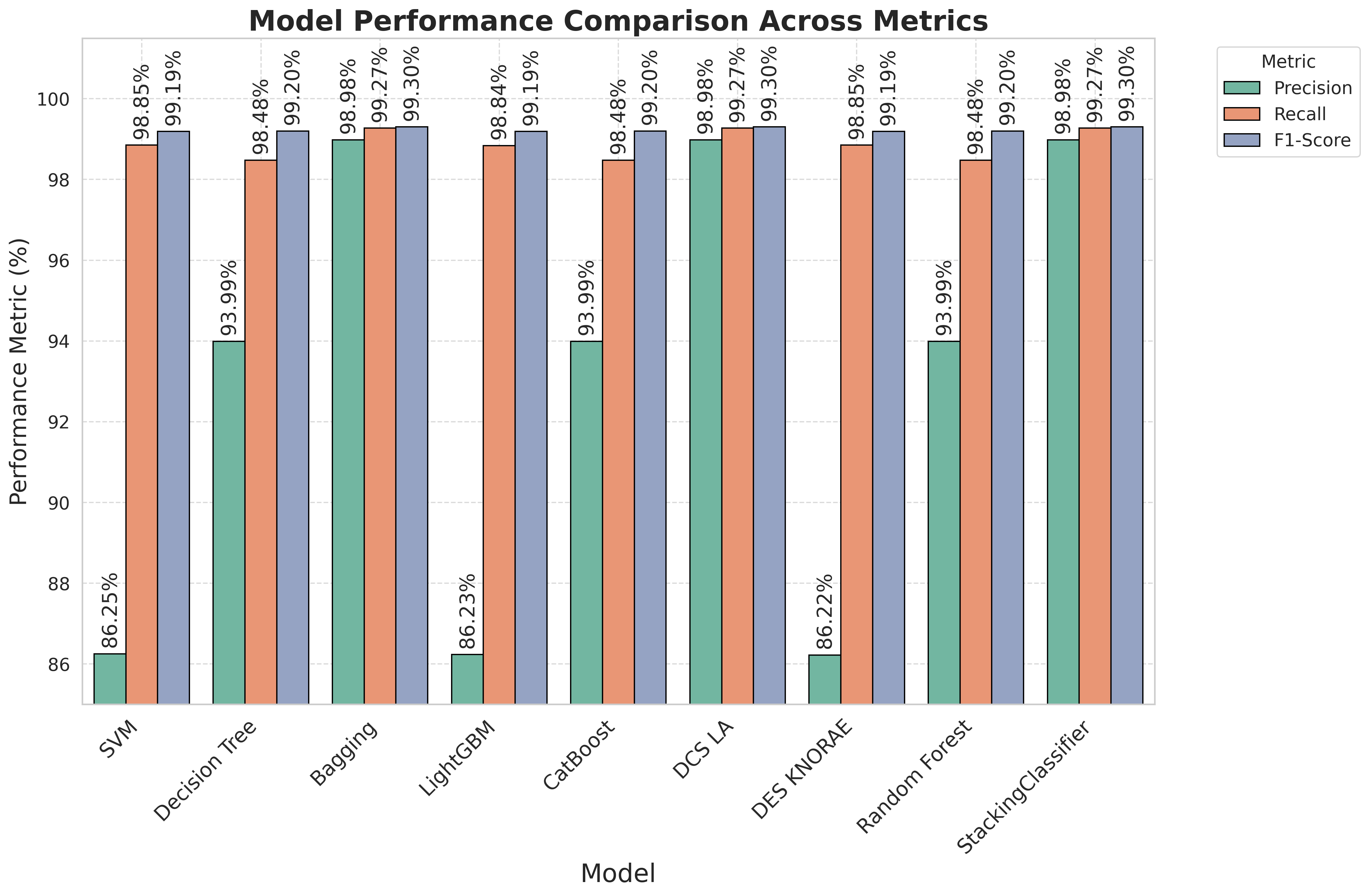}
\caption{Average Precision, Recall, and F1 Scores for Each Model Across Both Classes.}
\label{fig:model_comparison_plot}
\end{figure}
The Support Vector Machine (SVM) model showed improved results after applying SMOTE, but it still fell short in precision and recall compared to the ensemble-based methods. In contrast, LightGBM and CatBoost achieved notable results, particularly in the Down class, demonstrating F1-scores of 0.9396 and 0.9848, respectively. Figure~\ref{fig:model_comparison_plot} shows the average precision, recall, and F1 score of each model, giving an overall view of the performance of the models with respect to classification across the in-service versus out-of-service classes. Since all these metrics are averaged here, this figure can easily compare model reliability and consistency, allowing a balanced interpretation of each model's overall accuracy.

\subsection{Analysis of SMOTE's Impact}

The application of SMOTE greatly improves classification with the use of this highly imbalanced dataset, especially for those models that were quite poor in detecting out-of-service instances. That is quite clearly visible in the case of the SVM model, for which the average F1-score was around 0.50 before the application of SMOTE—hence showing its inability to find instances from a minority class effectively. After applying SMOTE, the average F1-score of the SVM has risen to 0.86, indicating a great improvement in its capability of classification without big bias over either class.

Similarly, Random Forest and LightGBM showed remarkable improvements in average F1-scores after the application of SMOTE. The F1-score for Random Forest rose from 0.78 to 0.99, showcasing its strong adaptability to the resampled dataset. In its turn, LightGBM showed a great boost: its average F1-score improved from 0.73 to 0.94, reiterating how important this step was in enhancing LightGBM sensitivity toward out-of-service instances.

\begin{figure}[b!]
\centering
\includegraphics[width=0.85\textwidth]{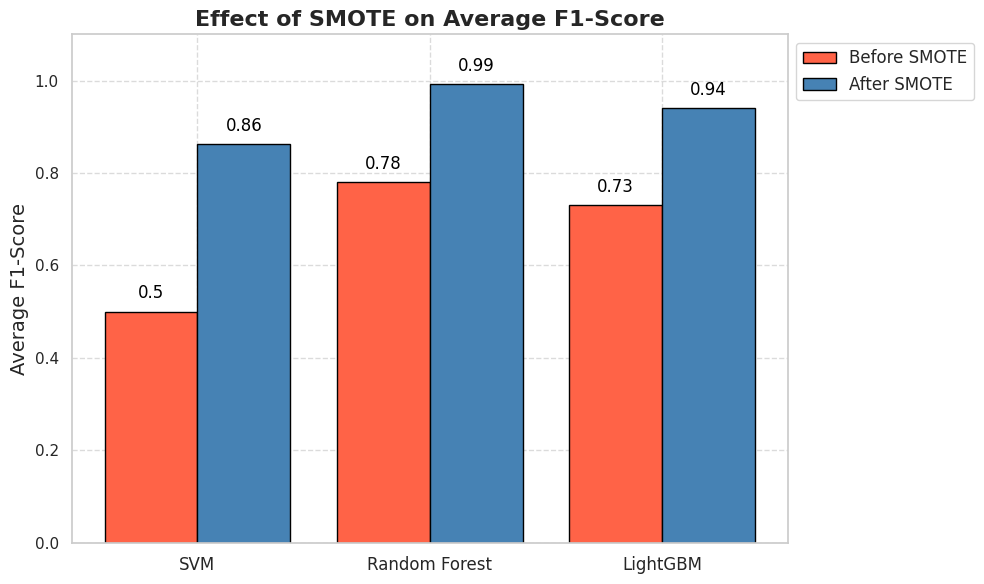}
\caption{Average F1-scores for SVM, Random Forest, and LightGBM Before and After SMOTE.}
\label{fig:comparison_plot}
\end{figure}

Figure~\ref{fig:comparison_plot} illustrates the effect of SMOTE on the average F1-scores of SVM, Random Forest, and LightGBM, indicating where it significantly improved after the application of SMOTE in either class.

\subsection{Discussion}

The results reveal that ensemble methods such as Random Forest, Bagging, and DCS LA consistently outperformed simpler models like SVM and Decision Tree. These ensemble models leveraged their ability to combine predictions from multiple weak learners, resulting in higher accuracy and robustness for both in-service and out-of-service classifications. Additionally, the application of SMOTE was pivotal in enhancing recall for the minority class (Down), ensuring that the models could reliably detect out-of-service ATMs.

The integration of SMOTE with ensemble learning techniques delivered the best results, with Random Forest standing out as the top performer, achieving the highest F1-scores overall.

\section{Conclusion}

The present research employed various machine learning models with the aim of improving the reliability of ATM networks by precisely detecting out-of-service statuses and reducing false alarms. Among them, the stacking classifier showed the highest overall accuracy at 99.29\%, outperforming the individual models and other ensemble methods.

A stacking classifier with Random Forest, LightGBM, and CatBoost could effectively combine their strengths through a Logistic Regression meta-learner to capture the patterns of intricacy in the data and yield a more balanced prediction. This achieved significant reductions in both false alarms and missed alarms, significantly improving operational efficiency in ATM networks.

It is very important to provide the implementation of SMOTE in view of the class imbalance problem, which allows the model to learn properly about the frequent and rare events.

The ramifications of such findings on financial institutions would, therefore, be considerable, offering a practical means of enhancing the dependability of ATM networks, cutting superfluous maintenance costs, and advancing customer satisfaction. Future work could explore integrating real-time monitoring techniques or applying the proposed methodology to other reliability-critical systems, such as healthcare or transportation networks.

\section*{CRediT authorship contribution statement}

\textbf{Alireza Safarzadeh}: Conceptualization, Methodology, Software, Data Curation, Formal Analysis, Investigation, Visualization, Writing – Original Draft.

\textbf{Mohammad Reza Jamali}: Technical Supervision, Resources, Validation, Writing – Review \& Editing.

\textbf{Behzad Moshiri}: Supervision, Project Administration, Funding Acquisition, Writing – Review \& Editing.

\section*{Declaration of competing interest}

The authors affirm that this research was funded by Pulseware Company. There are no other competing financial interests or personal relationships that could have influenced the work reported in this paper.

\section*{Data availability}

The data supporting the findings of this study are proprietary to the collaborating banks and were accessed and analyzed under strict confidentiality agreements. Due to privacy and contractual obligations, the data cannot be made publicly available or shared.

\section*{Funding}

This research was funded and supported by Pulseware Company.

\section*{Compliance with ethical standards}

This article does not involve any studies with human participants or animals conducted by any of the authors.

\bibliography{mybibfile}
\end{document}